\documentclass{article}
\usepackage[utf8]{inputenc}
\usepackage{authblk}
\usepackage{graphicx}

\usepackage{setspace}

\usepackage{biblatex}
\title{Advancing Community Engaged Approaches to Identifying Structural Drivers of Racial Bias in Health Diagnostic Algorithms}

\author[1]{Jill A. Kuhlberg}
\author[2]{Irene Headen}
\author[3]{Ellis A. Ballard}
\author[4]{Donald Martin, Jr.}
\affil[1]{System Stars, LLC}
\affil[2]{Drexel University}
\affil[3]{Washington University in St. Louis}
\affil[4]{Google}
\date{June 2020}

\begin{document}

\maketitle
\begin{abstract}
    Much attention and concern has been raised recently about bias and the use of machine learning algorithms in healthcare, especially as it relates to perpetuating racial discrimination and health disparities. Following an initial system dynamics workshop at the Data for Black Lives II conference hosted at MIT in January of 2019, a group of conference participants interested in building capabilities to use system dynamics to understand complex societal issues convened monthly to explore issues related to racial bias in AI and implications for health disparities through qualitative and simulation modeling.  In this paper we present results and insights from the modeling process and highlight the importance of centering the discussion of data and healthcare on people and their experiences with healthcare and science, and recognizing the societal context where the algorithm is operating. Collective memory of community trauma, through deaths attributed to poor healthcare, and negative experiences with healthcare are endogenous drivers of seeking treatment and experiencing effective care, which impact the availability and quality of data for algorithms. These drivers have drastically disparate initial conditions for different racial groups and point to limited impact of focusing solely on improving diagnostic algorithms for achieving better health outcomes for some groups.  
\end{abstract}

\spacing{1.5}
\section*{Background \& Motivation}

\subsection*{Problems with ML/AI-based Products in Health}
Recent news stories have brought public attention to the real and potential impact of biased algorithms used in high-stakes machine learning (ML)/artificial intelligence (AI)-based decision support systems \cite{challen_artificial_2019, oh_diversity_2015}. 
%The reasons underlying this systematic bias vary, but all similarly relate to a critical lack of Black and minority %representation across the artificial intelligence space. In particular, many stories emphasize how disparities in the %amount of algorithm training data available for different groups can create biases in artificial intelligence (AI) based %algorithms' performance that are especially harmful for marginalized groups, including people of color %\cite{khullar2019disparities,kirchner2015discrimination}. 
The reasons underlying these problematic outcomes vary, but many stories emphasize how disparities in the amount of algorithm training data available for different groups can lead to biases in ML algorithms that are especially harmful for marginalized groups, including people of color \cite{khullar2019disparities,kirchner2015discrimination}.

Some stories highlight how problematic proxies used in a machine-learning (ML) system can perpetuate historical racial discrimination \cite{buolamwini2018gender,demooy2017bias,ingersoll2019algorithm,martin2020participatory}. Still, others point to the lack of diversity in the teams and organizations involved in the ML/AI product development process resulting in biased assumptions being reflected in product design and performance. All of these issues emphasize critical areas of improvement needed to create racially equitable ML/AI-based supportive technologies, and more complete consideration of the wider context of structural inequity, especially relating to feedback structures of trust and distrust arising from historical legacies of racial inequity \cite{powell2008racism}, may be required to fully achieve this goal.   

System dynamics is specifically situated to provide a lens to identify underlying tangible and intangible accumulations, feedback structures and implicit and explicit decision rules that may perpetuate historical racial discrimination. Through qualitative causal maps and through formal, quantified simulation models, SD provides a language to formally represent causal theories to be tested, refined, and improved \cite{richardson2011reflections}.  While a few practitioners have started a discussion about the use of SD to address issues related to racism \cite{frerichs2016}, these approaches have been largely expert driven and have focused on the consequences of racism rather than the mechanisms through which historical injustices have embedded racial bias in our societal institutions. A central conundrum of pursuing a formal SD modeling approach to examine drivers and consequences of structural biases being reflected in data is that the empirical quantitative data through which one might validate or build confidence in model structure is influenced by the same structural bias that we are seeking to explore. By pursuing a conventional, expert-driven or quantitative data-driven modeling approach, we risk being caught in a circular loop in which the underlying structure that we seek is obscured by the structure itself.

Community based system dynamics (CBSD) provides an approach to engage with experts and non-experts who are directly affected by problems of algorithmic bias to better inform SD model formulation, while also building understanding, insight, and motivation to act against these structures.  CBSD is a participatory approach to system dynamics modeling that centers empowerment, mobilization, and building communities of practice \cite{hovmand2014}. Community participation in CBSD is a process,  to reference Forrester, to access the "mental database" of people embedded in systems \cite{forrester1992data}. However it is not limited to engaging participants as informants to derive or validate model structure.  Rather, it starts with capacity building for understanding the underlying principles of system dynamics as they manifest in participants' lives.  From this foundation of trust and insight, model development is an investment in developing capabilities within communities to address the proximate problem that motivated the modeling effort, but also to embed systems thinking and SD modeling capabilities that may persist after the modeling project is completed.  This dual approach to SD model formulation for insight and development of capabilities within communities suggests an approach that increases the likelihood that steps taken to eliminate racial bias in AI are sustainable.

This paper describes a model developed to generate insights about how the development of ML/AI-based products used to aid in diagnosis reflect and impact racial health disparities. Our community based system dynamics approach to the modeling process facilitated a deep discussion about the feedback structures linking the data collection and algorithm training procedures to historical memory of injustices caused by healthcare and medical research fields.  Policies tested highlight the challenges and delays in the collection of more data for algorithm training for underrepresented racial groups in the current structure of the system, and the limits of focusing on algorithmic improvement alone. We close with a discussion of future directions for modeling in this space.  

\section*{Project Design \& Modeling Approach}
\subsection*{Project Background}
The modeling presented in this paper occurred between September 2019 and January of 2020, and was embedded in a larger effort to engage participants from the Data for Black Lives II (D4BL II) Conference (held in January 2019) interested in applying system dynamics to understand and address complex problems impacting and involving Black Americans.  

At D4BL II, seventy conference participants attended a system dynamics workshop, where 10 community facilitators led small subgroups of participants in group model building activities to explore topics related to the expanding racial wealth gap in the US. Interested participants could attend a follow-up session the following day, which gave space for a deeper discussion around the origins of SD, participatory approaches to SD, additional qualitative modeling practice, and exposure to simulation. Based on discussions from the community of participants and facilitators of the workshop and follow-up session, a series of informal monthly calls were proposed to use SD to explore specific issues relevant to D4BL II participants, and to use the process to build SD capabilities among participants to use SD in their own work. The specific issue of racial bias in AI/ML algorithms used in healthcare and medical technologies was chosen as the substantive focus for the calls.  

\subsection*{Participants}  The calls were co-organized by three individuals: one Black woman, one White woman, and one White man.  One call organizer was also a facilitator during the D4BL II SD workshop, where the other two participated as process coaches and facilitated the follow-up SD session. At the start of the calls, one organizer had been exposed to SD for less than one year, and other other two had been practicing and teaching SD for several years.  One facilitator was a substantive expert in racial health disparities. 

The calls were advertised via email to the list of interested participants from the D4BL II SD workshop.  Interested individuals responded with availability via survey, and interested individuals were emailed monthly invitations for video conference meetings. In addition to the call organizers, a small group of participants (ranging from 2-6) joined each call. While one participant joined every call, other participants joined between  1-3 calls for a total of 8 separate individuals participating at some point during the calls.  Participants worked in education, environmental justice,  youth development, government, public health  and tech.  Three participants identified as female, two as male, one as gender non-binary and two did not specify.\footnote{Demographic information on the race of the participants was not collected.} Participants joined from eastern, central, and pacific time zones.

\subsection*{Modeling Process \& Relevant Components}
All modeling occurred virtually through video conference calls, as no two participants were geographically co-located. Three types of calls facilitated the modeling process: 

\begin{itemize}
    \item \textit{Community Calls}  The 90 minute community calls aimed to deepen and advance model-based discussions around the issue of data and algorithmic bias in health and use the example to introduce SD modeling concepts (i.e., reinforcing and balancing feedback, accumulation, simulation, etc.). These calls were held monthly, recorded and shared with individuals who expressed interest in the call, but were unable to join.
    \item \textit{Planning Calls} In addition the community calls, the call organizers held 30-45 minute planning calls to establish goals and content required for the upcoming call, reflect on issues from the previous call, and review offline communications via the call's accompanying discussion board. There were about two planning calls for each community call. 
    \item \textit{Team Modeling Calls}  The goal of these 90 minute modeling calls were for the call organizers to work exclusively on iterating on the model as a small group, using call discussions and discussion board activity as starting points.  These calls involved electing a ``driver'' for the modeling, and that person sharing the software's interface through screen-sharing as the group talked through and built the model in real-time together. These calls were often recorded. 
\end{itemize} 
\subsubsection*{Community Call Roadmap}
The community call roadmap was not set prior to the start of calls to allow for flexibility in the learning process as well as community feedback as the calls evolved. The initial community call was the only call that used a preset agenda, informed by survey feedback and D4BL II workshop debrief, in order to seed community participation and engagement. The initial call introduced the substantive topic anchoring the subsequent calls, and reviewed causal loop diagramming tools that participants had been exposed to in the D4BL II workshop and follow-up session. An example causal loop diagram (CLD) was generated in the spirit of some of the relevant news stories related to racial biases in AI/ML based products in healthcare systems. The CLD presented was used to both review skills learned in the D4BL II workshop and provoke discussion and critiques that would be incorporated into the next model iteration shared on the call. The subsequent call introduced accumulations and stock and flow diagramming conventions with the second version of the model that integrated discussion from the previous call and discussions from the offline engagement platform, and was again used to deepen and refine the discussion around the topic. On the third call, a revised and simulating stock and flow model was shared, informed by participant and organizer discussion on the second call, from the offline engagement platform, and through the team-based model building. Organizers on the fourth and final call presented another iteration of the simulation model, tested a set of policies, and discussed the implications. 

\section*{Results}

\subsection*{Reference Mode}
\begin{figure}[ht]
\caption{Initial reference mode (A), and revised reference mode following community discussion(B)}
\centering
\includegraphics[width=1.1\textwidth]{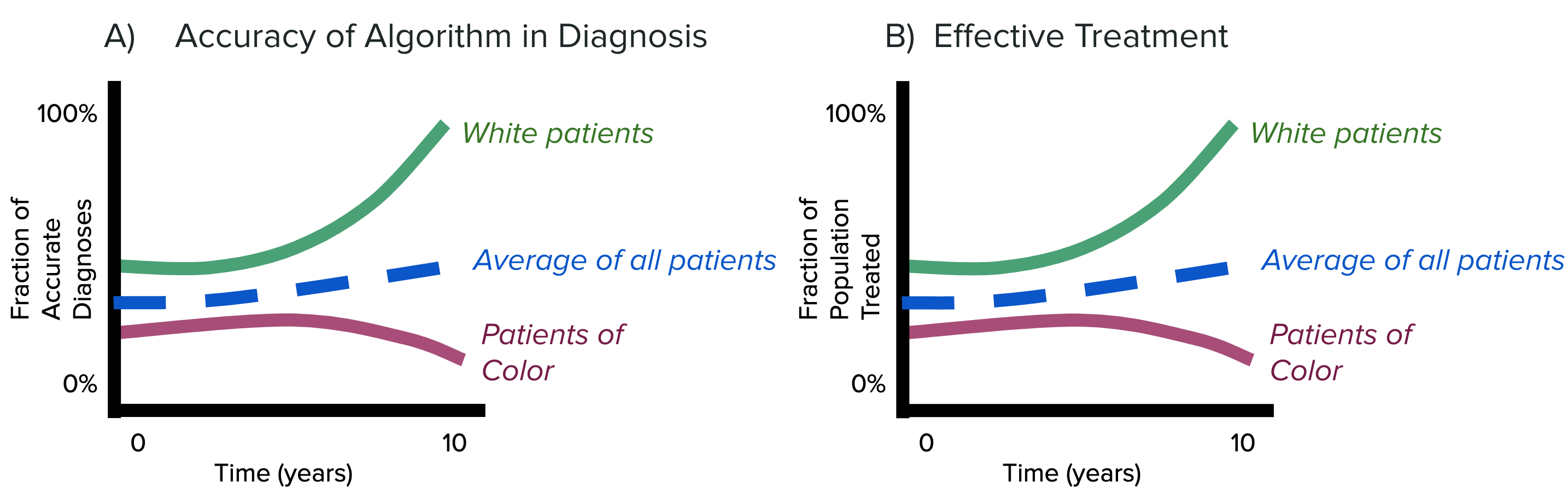}
\label{fig:InitialRefMode}
\end{figure}

As a taking off point, the organizing team chose an initial reference mode for the modeling effort that reflected the mainstream news framing of several health algorithm related issues: accuracy of the algorithm in detecting an underlying health condition (See Figure \ref{fig:InitialRefMode}).  Over the course of the first call, participants chose to shift the reference mode to the fraction of the population developing a condition who obtain effective treatment, a shift which moved to center the modeling effort to focus on the lives of those experiencing the illness and treatment. Call participants also discussed how illness, treatment and decisions around which illnesses are allocated resources to improve treatment are constructed within a racially biased context. Instead of modeling a particular health condition, the group decided to use the modeling process to explore more general feedback structures involved in algorithm development, disease progression, and treatment. This both allowed for broader engagement of community participants as well as seeded rich discussions of how model structures might differ for different disease outcomes that have different impacts on people of color.  

\subsection*{Model Structure}
\begin{figure}[ht]
\caption{Causal loop diagram of the most relevant feedback structures in the simulation model around improving treatment through ML/AI algorithm assisted technologies. Policies tested are noted in \textit{\underline{underlined and italicized text}} and dashed-line causal linkages.}
\centering
\includegraphics[width=1.1\textwidth]{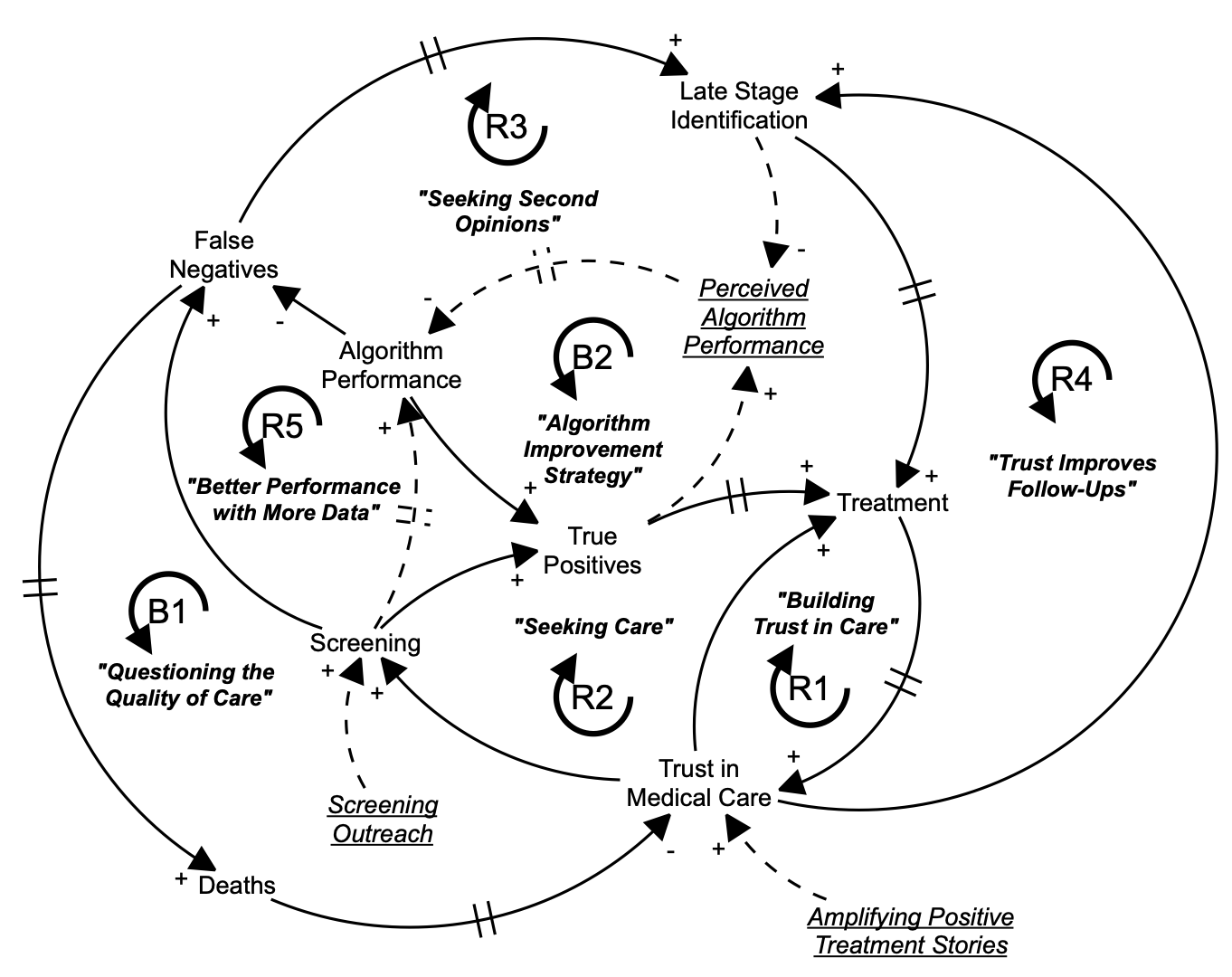}
\label{fig:CLD}
\end{figure}

Figure \ref{fig:CLD} provides the most relevant feedback loops in the simulation model focused on improving treatment rates using AI/ML assisted diagnosis technologies produced over the course of the calls. 

Four key reinforcing loops linked trust in medical care and screening and treatment (at early and late stages of condition development).  Trust in medical care increases the rate at which individuals seek and receive treatment, and positive or effective treatment increase trust in medical care (R1), all things being equal. Trust built from positive experiences with treatment increase other engagement with medical care including seeking care for regular check ups or screenings, which can uncover health conditions that can benefit from treatment, increasing the positive experiences with medical care, further building trust (R2). Trust in medical care can also increase the probability that individuals will seek out a second opinion after a misdiagnosis due to an undetected condition in the screening process, which if detected in the follow-up appointments, can increase referrals to treatment, and increase trust in medical care (R4). It is important to note the delays that the group identified, which could be categorized into three main types: treatment delays (from screening to effective treatment), delays related to updating perceptions and memory of historical group experiences with medical care, and delays in acquiring training data and employing an ML/AI based system to assist in diagnosis.

 A balancing loop (B1: \textit{Questioning the Quality of Care}) limited the positive gains of those reinforcing loops, as increased screenings would also lead to more false negatives \textit{ceteris paribus}, which would lead to more deaths, reducing trust in medical care, further reducing screening rates. This loop interacts with the other reinforcing loops as its impact on trust also affects treatment seeking.  Leverage points discussed by the group are also mapped onto the CLD and will be described in a the Policy Tests section.

The simulation model was organized in three main sectors: screening and treatment, algorithm development and diagnostic performance, and a trust and memory sector described in more detail in this section. The model was built using Stella Architect 1.9, and full model documentation can be shared by contacting the authors. The boundaries of the model (e.g., what is endogenous, exogenous, and excluded) are described in Table \ref{table:BoundaryChart}.

\begin{table}[ht]
\begin{tabular}{llll}  
\cline{1-3}
\multicolumn{1}{|l|}{\textbf{Endogenous}}             & \multicolumn{1}{l|}{\textbf{Exogenous}}           & \multicolumn{1}{l|}{\textbf{Excluded}}    &  \\ \cline{1-3}
\multicolumn{1}{|l|}{People screened}                 & \multicolumn{1}{l|}{Availability of treatment}    & \multicolumn{1}{l|}{Quality of treatment} &  \\
\multicolumn{1}{|l|}{People treated}                  & \multicolumn{1}{l|}{Insurance coverage}           & \multicolumn{1}{l|}{False positives}      &  \\
\multicolumn{1}{|l|}{Deaths from condition} & \multicolumn{1}{l|}{Mortality rate for condition} & \multicolumn{1}{l|}{}                     &  \\
\multicolumn{1}{|l|}{Algorithm performance}           & \multicolumn{1}{l|}{Time to forget deaths}        & \multicolumn{1}{l|}{}                     &  \\
\multicolumn{1}{|l|}{Memory of deaths}                & \multicolumn{1}{l|}{}                             & \multicolumn{1}{l|}{}                     &  \\
\multicolumn{1}{|l|}{Trust in healthcare}             & \multicolumn{1}{l|}{}                             & \multicolumn{1}{l|}{}                     &  \\ \cline{1-3}                     & 
\end{tabular}
\caption{Model boundary table defining what is endogenous, exogenous, and excluded from the simulation model.}
\label{table:BoundaryChart}
\end{table}

\subsubsection*{Screening \& treatment}
\begin{figure}[ht]
\caption{Screening and Treatment Stock Flow Structure}
\centering
\includegraphics[width=1\textwidth]{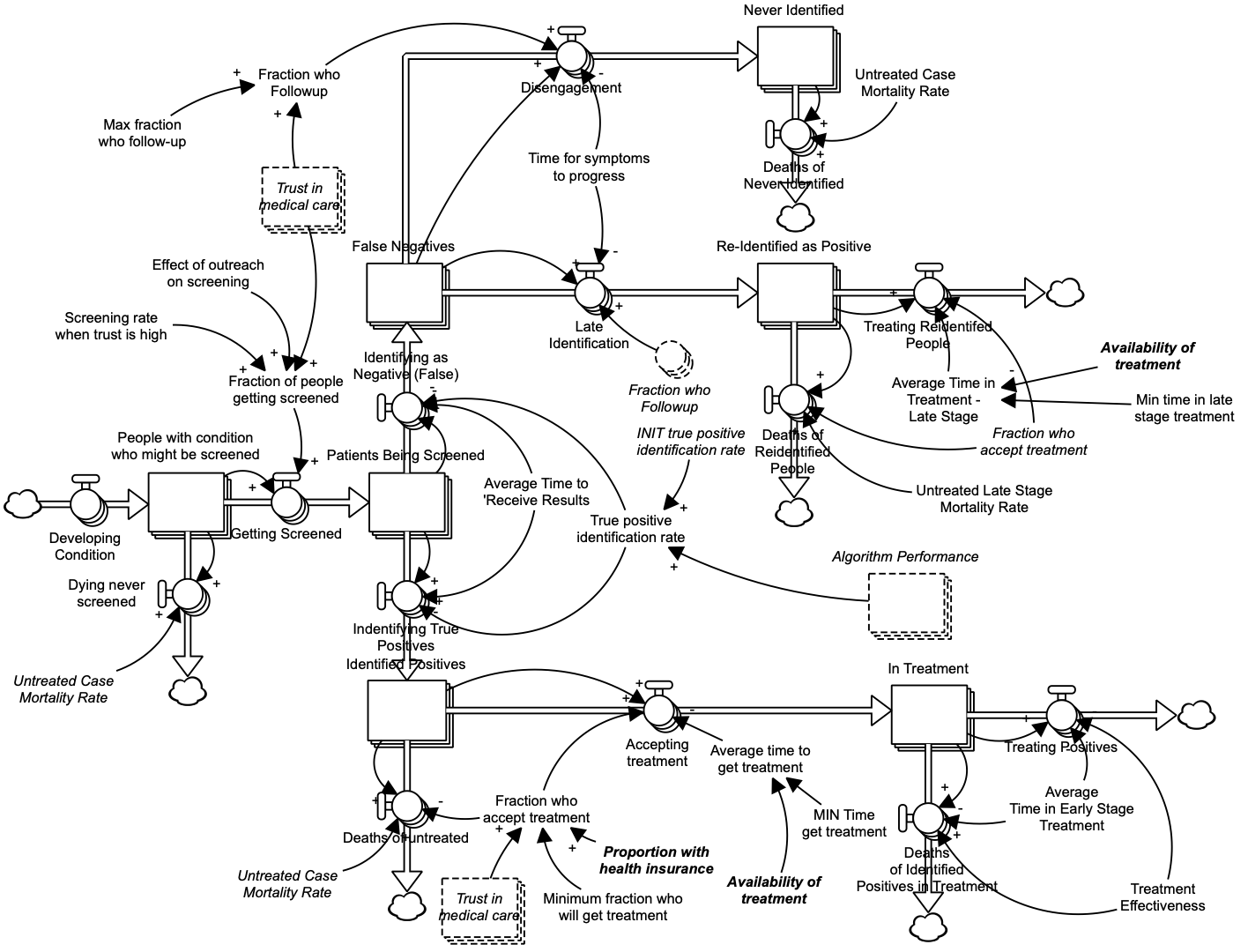}
\label{fig:TreatmentScreening}
\end{figure}

The model simulates the movement of a population through the healthcare system beginning with the early development of symptoms of a particular condition like cancer.  Figure \ref{fig:TreatmentScreening} shows the stocks in the screening and treatment sector. The people with the condition who might be screened represents the number of people in the population (or of a particular subgroup, if the model is arrayed/sub-scripted) who have developed the condition, but have yet to be screened by a medical professional for it. The stock is added to from the flow of people developing the condition each year, and people leave the stock either because they have sought medical care that involved screening, or because they die from the condition before being formally screened for it\footnote{Deaths before screening and all deaths in the current model are assumed to be attributable to the condition.} The stock of people being screened are those who are awaiting the screening appointment and results of the diagnostic screening, and move to one of two diagnosis stocks upon receiving screening results, true positives or false negatives, as function of the time it takes to receive screening results and true positive rate of the AI/ML-based diagnostic tool. The explicit delineation of false positives and false negatives within the model structure was a product of offline participant engagement with the CLD presented in community call 1. The participants observation that, especially in the context of how racial bias may arise in algorithm based diagnosis, false positives and false negatives may not have the same impact. This led the call organizers to specifically revise the model to include this distinction during a modeling call between community calls 1 and 2.

Following the path of those who receive a true positive diagnosis, the model reflects the delay upon receiving a diagnosis and the decision to pursue treatment. The number of people who begin treatment is a function of the average trust in medical care, health insurance coverage, and the availability of actual treatment procedures, which impacts the time it takes to receive treatment upon seeking it.  Higher delays to access treatment, and low rates of health insurance coverage and trust in medical care, increase the number of people who die each year after having received a positive diagnosis.  The in treatment stock includes all of those who have started treatment.  The rate of receiving effective treatment is a function of the length of treatment and probability of treatment success, and the rate of death in treatment is a function of time in treatment and 1 minus the probability of treatment success.

Returning to the stock of people being screened and following the path of those wrongly given a negative diagnosis, the model reflects that those individuals enter a stock where their symptoms continue to progress, without treatment. As symptoms become more noticeable, people in this stock may question the results of their initial screening and seek treatment, or may question the competency of the medical system and not follow-up.  The fraction of those who follow up after symptoms increase in severity in this model is a function of trust in medical care. Similar to those that entered treatment upon earlier detection, those that enter treatment at this later stage can either be treated effectively, or die while undergoing treatment. The probability of successful treatment at late stage is lower than early stage treatment, since the condition is considered to be more severe at the later stages.

\subsubsection*{Trust \& memory}
\begin{figure}[ht]
\caption{Stocks and flows of trust and memories related to experiences with medical care.}
\centering
\includegraphics[width=1\textwidth]{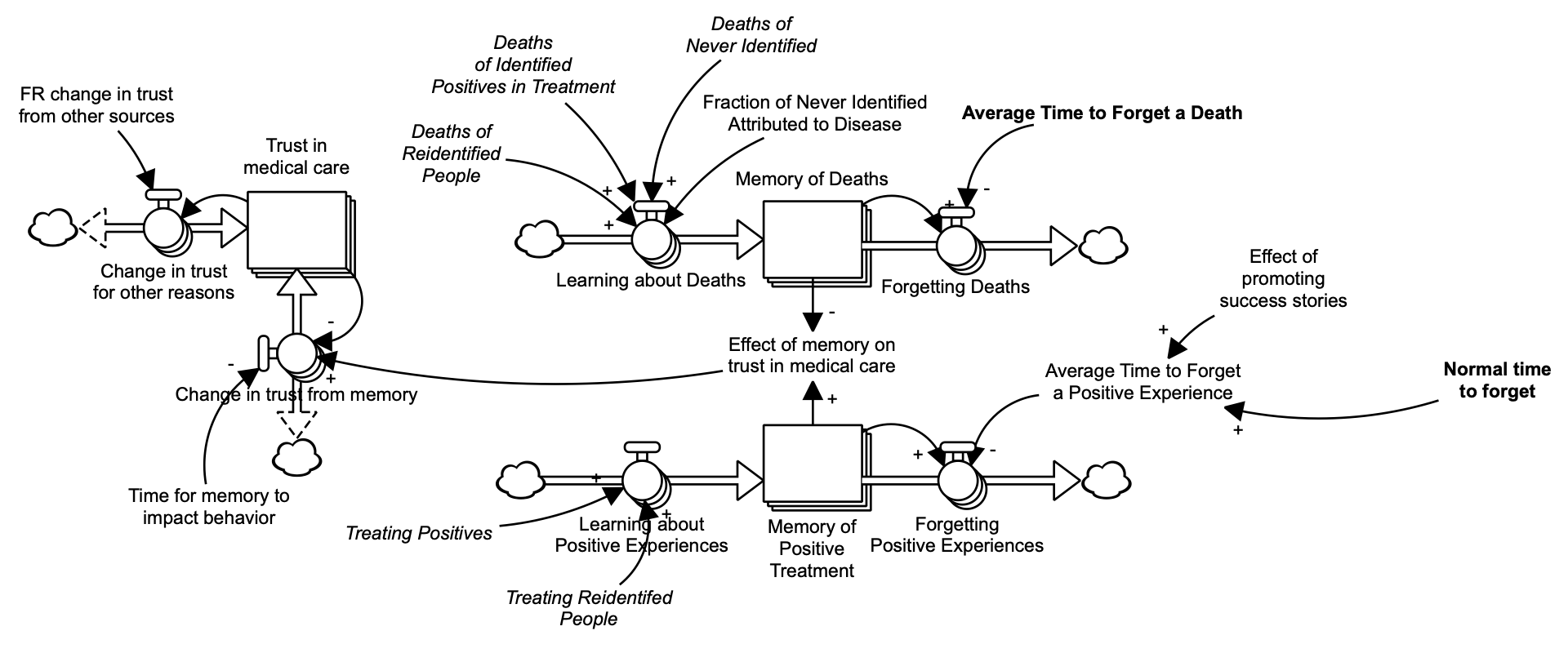}
\label{fig:TrustMemory}
\end{figure}
Over the course of the development of the screening and treatment sector, important community conversations came up about how many of the stocks, such as effective treatment, and flows, such as time in treatment and fraction who follow up, systematically differ for Black and White Americans, revealing the need for a larger lens that extended far beyond the algorithmic performance in identifying true positives. In particular, an important set of narratives emerged throughout the calls about the endogenous formation of trust in medical care and the historical experience and memory of deaths attributed to poor quality or intentionally harmful medical care.  To reflect this structure, we first used two stocks to keep track of the memory of two types of experiences with medical care (See Figure \ref{fig:TrustMemory}).  One stock accumulated the memory of cases where medical care had a positive result (i.e., treatment was successful early or late) and the other accumulated the memory of cases where medical care had failed to intervene and prevent a death, and that death was attributed to the treatable condition. Both stocks had outflows for ``forgetting'', which operationally reduced the impact of each of those types of experiences on medical care over time.  The time delay to forget deaths was twice as long as the time delay to forget a positive experience. This estimation came from a discussion from personal experiences among participants that negative events have a more lasting impact than positive ones.  

A stock reflecting the average trust in medical care was impacted by the fraction of positive experiences with medical care over all experiences in current memory and an exogenous fractional rate of change in trust in medical care, which was added to account for all other sources of trust that were not a function of the deaths and treatment experiences related to the condition being tracked in the screening and treatment sector.

\subsubsection*{Algorithm development}
\begin{figure}[ht]
\caption{Stocks and flows related to training data and algorithm performance.}
\centering
\includegraphics[width=1\textwidth]{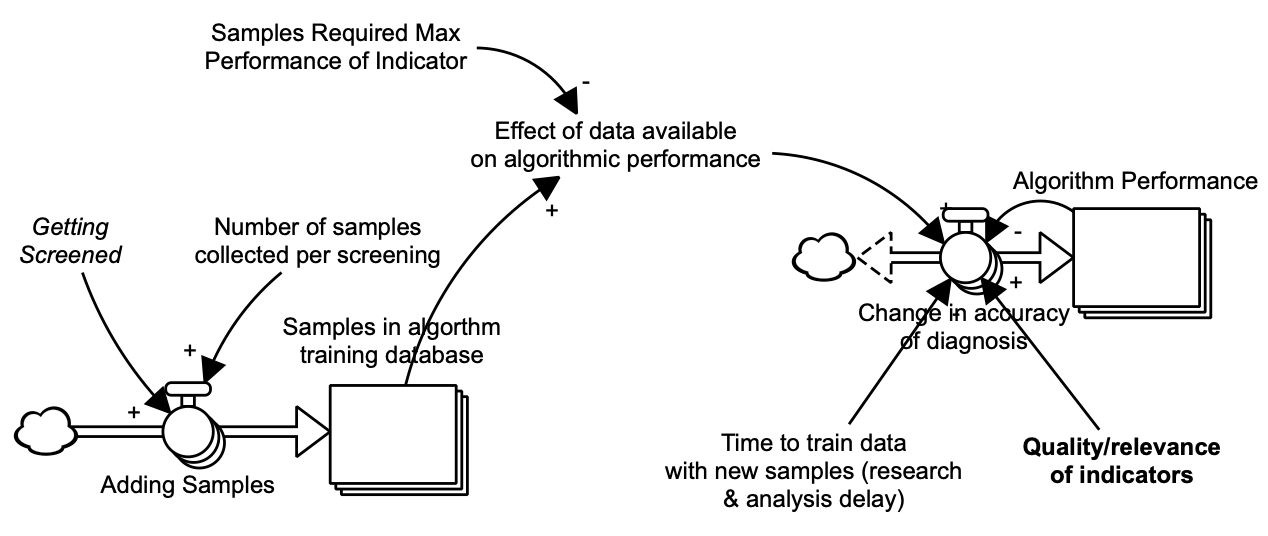}
\label{fig:AlgorithmSector}
\end{figure}

This model focused on the system structures that impact the collection of data that can be used to train diagnostic algorithms and the limits of their effectiveness.  The algorithm development sector features two main stocks (see Figure \ref{fig:AlgorithmSector}), one is the physical number of data samples collected that can be used for data training, that is increased by data collected in screening procedures (assuming patients allow their data to be used for this purpose). Another stock reflects the performance of the algorithm on a scale from 0 to 1, where 0 indicates a total inability to detect the presence of the condition in a true positive individual, and 1 indicates the best possible outcome, where the test is able to detect 100\% of all true positives.  An increase in samples available increases the algorithm's performance under the logic that more data allows for more nuanced understandings of the presentation of a condition. We also introduced a limit to the performance of the algorithm to reflect the use of technologies with biases and the collection of data on indicators that are biased in their ability to reliably detect the condition for all groups. 

\subsubsection*{Algorithm Development}

\subsection*{Model Testing}
\subsubsection*{Arraying the Model to Reflect Racial Disparities}
As the goal of the modeling project was to explore the impact and generation of racial biases in algorithm-assisted medical technologies, we arrayed the model for two groups: one with one set of parameters and initial conditions to reflect the experience of Black Americans and other for White Americans.

Estimates for parameter values for each group were informed by empirical data when available.  Model tests assumed that the number of new cases from each group per year was roughly proportional to the group's representation in the population. The proportion of the population with health insurance also varied across groups, with Black Americans insured at a lower rate than White Americans. Treatment was more available for White Americans than Black Americans, reflecting the known disparities related to the distance, availability of appointments, and cost (among other factors) related to beginning treatment \cite{ramirez2013time, peipins2013racial, foy2018disparities}. 

The model was initialized in equilibrium, since with the generic focus of the model we had no specific historical trend to replicate.  This entailed adjusting the initial values of the stocks in the screening and treatment sector to compensate for differences in the rate of developing the condition for the different groups, and for differences in the parameters discussed previously.  The initial values for the stocks of memory and trust were also adjusted to reflect the historical injustices of death and other negative experiences with the health care system of Black Americans, and the differential impact of the memory of these events on action for Black and White Americans. 

\subsubsection*{Policy Tests} 
The model simulations were run over a 20 year period, and policy tests were simulated as taking effect at the beginning of the fifth year in the simulation. Ideas for policy tests were the result of discussions on the community calls and in the modeling calls, as the modelers considered options that ranged from parameter adjustments to policies that changed the feedback structure of the system. They were additionally informed by input from public health practice on traditional approaches to behavior change, screening and testing interventions \cite{vancleave2012intvnts, cdcstrategies2016}. Each policy type is described below, and also represented in the CLD in Figure \ref{fig:CLD}, highlighted with dotted lines and italics.

\begin{figure}[ht]
\caption{\textit{Simulation results of the impacts of increased screening.}   An aggressive outreach strategy to increase screening by 15\% increases the fraction of Black Americans treated by only 2\%. The algorithm is perceived to be doing well until false negatives are discovered after a delay.}
\centering
\includegraphics[width=1\textwidth]{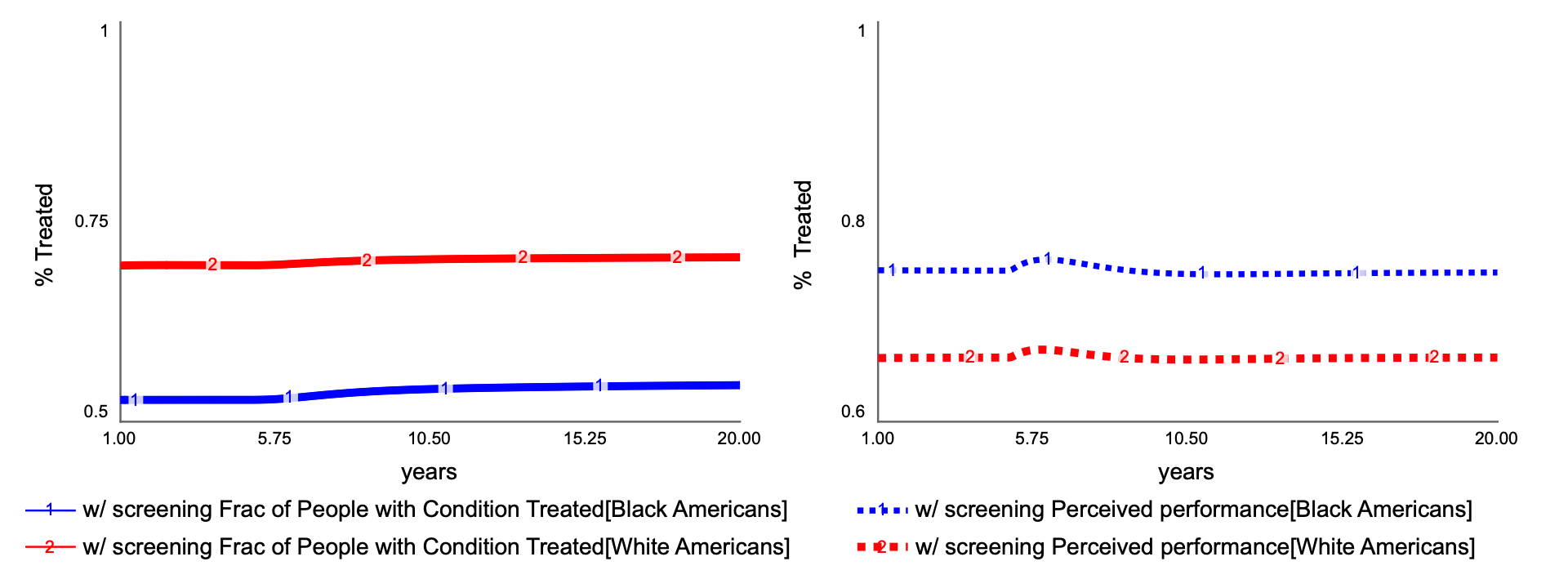}
\label{fig:ScreeningTest}
\end{figure}
\textit{Increasing screening} A frequent public health intervention suggested to increase diagnosis and treatment of a specific condition is to encourage screening \cite{vancleave2012intvnts}. In this model, this intervention is modeled as a step increase (of 15\%) in the fractional rate of people getting screened at time 5. Results of this policy are shown in Figure \ref{fig:ScreeningTest}. 

The aggressive outreach campaign to promote screening has only a marginal impact on the fraction of the population treated over time (gains of only about 2\%).  As nothing about the algorithm or access to treatment has changed in addition to the screening, more of the population gets screened for treatment, but several are misdiagnosed and their deaths after unsuccessful late stage treatment or having never been detected until death, add to the memory of deaths related to poor medical care, thus having no meaningful (positive) impact on trust in  medical care (Trust in  medical care increases  only  half a percentage point for Black Americans, and remains constant for White Americans).  Examination of the perceived performance (Figure \ref{fig:ScreeningTest}) of the algorithm shows that although the algorithm has not improved, it is perceived to increase, with the jump in positive diagnoses from increased screening. This perception is updated when false negatives are discovered as the symptoms of those initially screened as negative surface. By the end of the simulation, the perceived performance of the algorithm is slightly lower than initially thought, as more cases over time have shed light on the number of false negatives. (Note that algorithm  is  perceived to be functioning better for Black Americans, even though the actual performance is set lower for Black Americans. The lower rates of follow-up falsely inflate the algorithm's perceived performance.)

\textit{Amplifying positive experiences with medical care} Realizing that many individuals \textit{are} effectively treated, and that trust in medical care could be impacted by decreasing the memory of deaths and/or increasing the memory of positive experiences, we tested the impact of a public health marketing campaign to increase the visibility of positive experiences with medical care \cite{cdcstrategies2016}. We represented this strategy by a step increase in the time to forget a positive experience at year 5 (adding 3 additional years, to the ``normal'' 5 years).

\begin{figure}[ht]
\caption{\textit{Simulation results of the impacts of amplifying positive experiences with medical care.} The fraction of the population increases to a similar level as the promoting screening (\ref{fig:ScreeningTest}) over the 20 year time horizon, however, the positive story amplification strategy has more impact on trust in medical care, and has implications for the shape and future trajectory of treatment beyond 20 years.}
\centering
\includegraphics[width=1\textwidth]{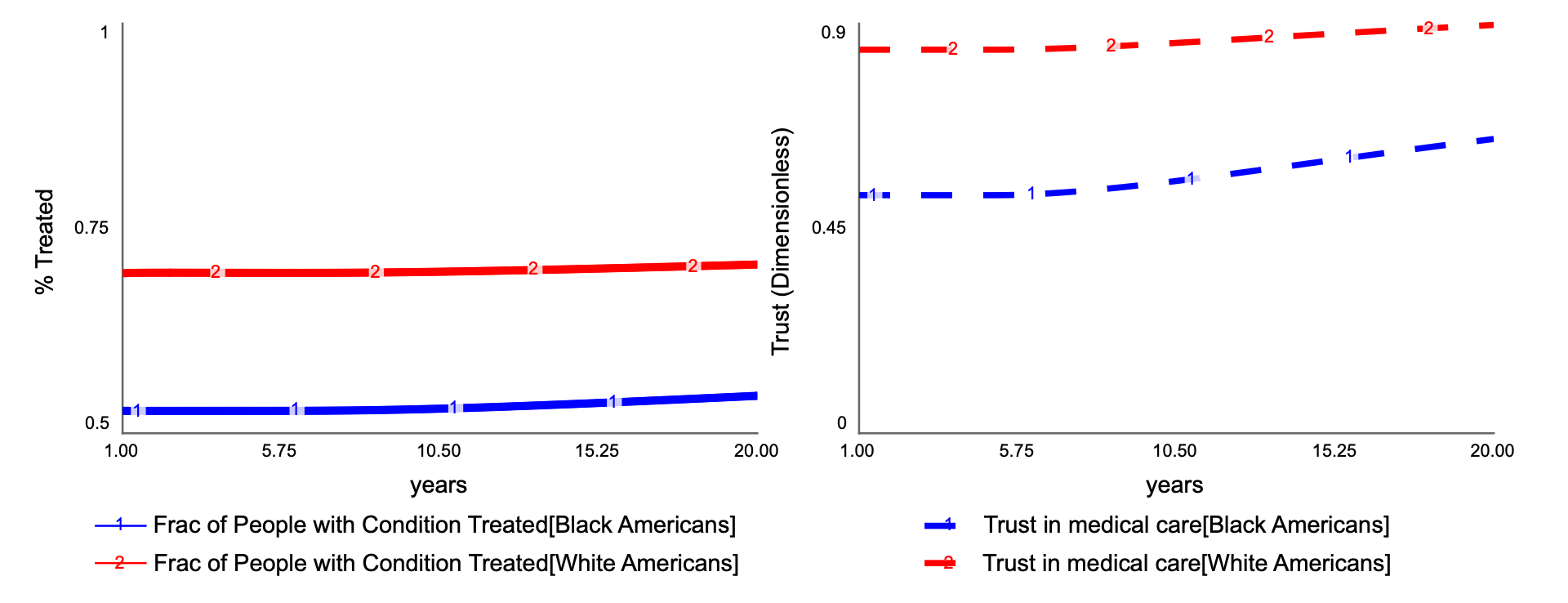}
\label{fig:HypePositive}
\end{figure}
The results of this strategy (Figure  \ref{fig:HypePositive}) are marginally better in terms of the fraction of individuals treated over the time horizon simulated, however, the real differences between this intervention and the screening outreach strategy appear more starkly when looking at its impact on trust in medical care. Trust increases much more dramatically, and that increase is what is behind the differences in qualitative shape of treatment outcomes between the  screening and positive hype strategies. In the former, the improvements in treatment outcomes appear to level off, and in the latter, be on an upward trend.  Simulating the model for an additional  20 years in the future (not shown) confirm that the positive hype strategy to have more of an impact (5\% more Black Americans treated for the condition).

\textit{Collecting more training data for the algorithm} While the algorithms used to inform decision support technologies are created at one point in time, one strategy would be more continuous collection of samples for the training data set. Adding training data collected at screening increases the algorithm's performance\footnote{The effectiveness of the algorithm is limited by the quality of the indicators chosen as markers for the condition and the ability for the current technologies to collect reliable data on those markers.}, reducing false negatives, deaths, increasing trust in medical care, which increases screening rates, and creates more opportunities to collect data to further refine the algorithm. We model this intervention that creates a reinforcing loop (R5 in Figure \ref{fig:CLD}), by allowing after 5 years of using the current algorithm, additional data to be collected each time someone is screened and added to the accumulated data from previous years/research studies. Data is collected until the average perceived performance across groups reaches an acceptable level of performance, creating an additional balancing loop (B2) in the model (see Figure \ref{fig:CLD}.

\begin{figure}[ht]
\caption{\textit{Simulation results of the impacts of improving the algorithm through increased data collection. Column [A] Shows results from using the average algorithm performance to inform data collection efforts, and column [B] shows how the using group-specific algorithm performance indicators impacts the percent of the population identified and treated.} }
\centering
\includegraphics[width=1\textwidth]{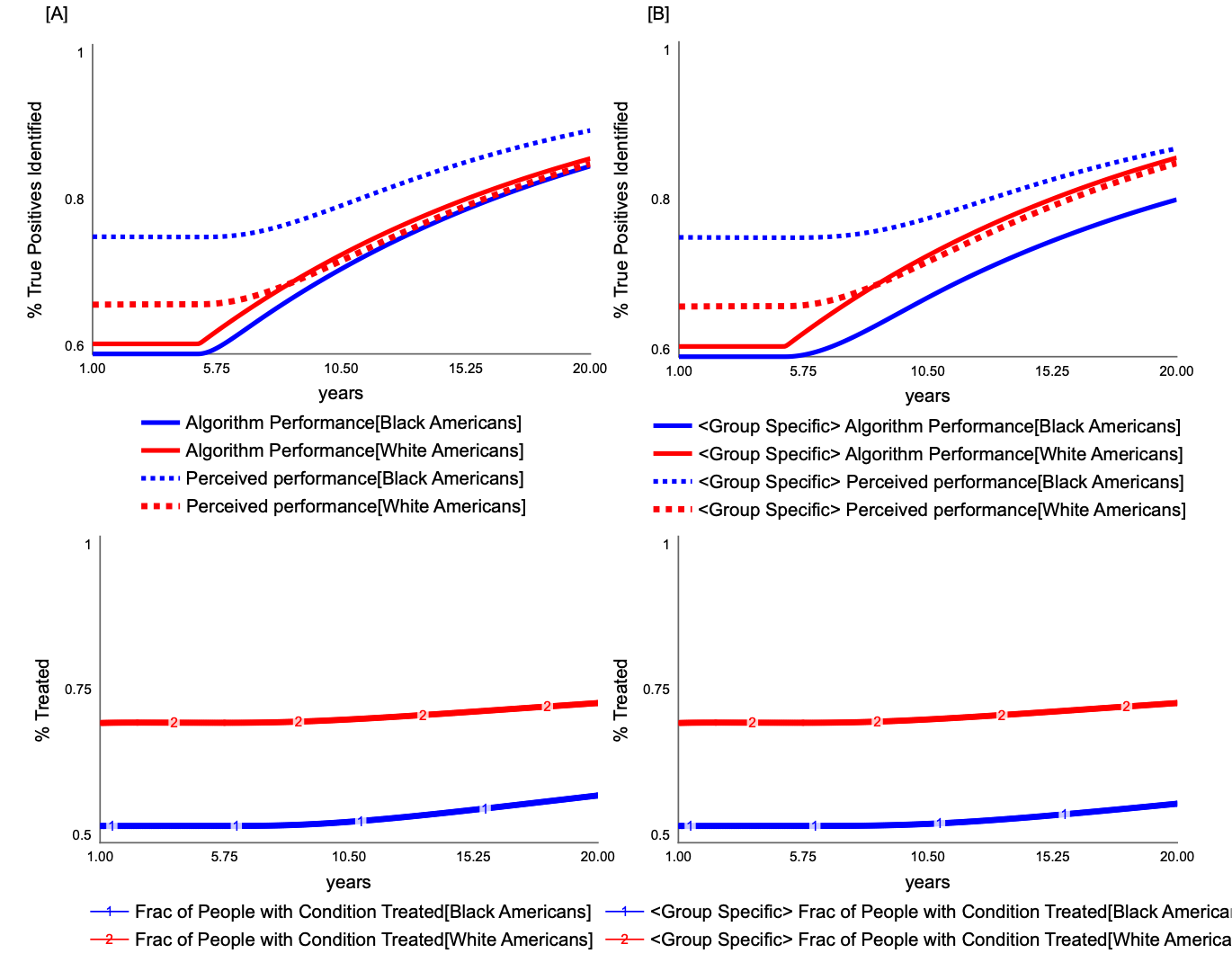}
\label{fig:AlgImp}
\end{figure}
The impact of a riff on this policy tests was also explored. In this version, data is collected for groups at different rates, as function of the diagnostic algorithm's performance for each group respectively. If the performance is perceived to be working at or above the acceptable threshold for one group, data collection will cease for that group, even if it is continuing for the other group's for which the diagnostic algorithm is performing worse.

Simulation results (Figure \ref{fig:AlgImp}) of this policy introduce a balancing feedback mechanism and both versions increase the fraction of individuals treated effectively, though the results of an improvement strategy  that uses the average perceived performance across groups (see Panel [A] in \ref{fig:AlgImp}) appears to have more impact than the group specific version (see Panel [B] of \ref{fig:AlgImp}) of this intervention.  This might be counter intuitive had previous simulation results not already shed light on the positive bias in perceived performance as result of lower follow up and treatment rates in Black Americans.  Since the diagnostic algorithm is perceived to be doing better for Black Americans, data collection and algorithm slows faster as it approaches what is considered acceptable performance, than White Americans. While both policies result in treatment rate improvements, both also show that the algorithm is performing better for White Americans than for Black Americans, while the perception is just the opposite. 

\section*{Discussion}
In this paper we describe a model developed through a rapid collaborative modeling effort, grounded in the principles of community based system dynamics.  The goals of developing this model were twofold. First, we sought to advance learning and dialogue around the mechanisms through which ML/AI-based approaches to healthcare delivery and decision making may perpetuate or exacerbate racial biases in health care system.  Additionally, we sought to use this platform as a tool to engage non-SD activist and academic communities to co-develop system dynamics capabilities through a project-based learning approach. Based on these dual goals, the resulting model is an artifact of a process; structural and process insights suggest substantive implications about our understanding of the structural drivers of bias in healthcare, and methodological implications around the use of project-based learning approaches to simulation development that may propose a path forward for future work in this space.

\subsection*{Structural Insights}

\subsubsection*{Model boundaries and reference mode} 
Early versions of this model focused narrowly on machine learning algorithm performance and data availability, in which available data was generated through sample collection from patients seeking treatment.  It became quickly apparent that this question was neither structurally interesting-- the model basically consisted of an algorithm training balancing loop and healthcare screening trust reinforcing loop-- nor substantively interesting to members of the community call. The essential question after early iterations of this model was, ''Where are the people?'' If we are interested in algorithm performance because of its influence on human health and racial disparities, the boundaries of the model needed to reflect that goal.  In later versions of the model the reference mode has shifted, focusing on healthcare treatment as the problem of interest as a way to integrate the operational mechanisms of AI performance improvement and contextualize that algorithm performance within healthcare seeking behavior. 

However, though this model focuses on treatment as the problem of interest, we could also expand the boundary of the model to examine the role of diagnosis algorithms in promoting prevention of the development of cases.  Alternatively, we could consider morbidity or mortality itself as our problem of interest, examining the interaction of diagnosis or treatment planning algorithms as a lens to examine disparities in health outcomes rather than health treatment outcomes.

\subsubsection*{Current and future use of diagnosis algorithms} In modeling the hypothesized mechanisms through which data is generated and feeds back into machine learning algorithm training, it became apparent that a feedback theory of algorithm training is optimistic at best.  It is an open question, from our perspective, whether algorithm performance is meaningfully assessed by the developers and deployers of those algorithms, and how those developers respond to positive or negative performance.  The structure presented in this model is a hypothesis of the set of information cues and policy decisions that could be undertaken in a ideal world where algorithm developers have a connection to the implementation of those algorithms, and have incentives to take steps to improve performance.   A more realistic assumption is that algorithms may be developed up to some fixed performance standard, and then are utilized until a better algorithm replaces it.  However, the initial results of this model suggest that even in an ideal scenario in which algorithm development exists within a learning cycle and incentive structure to improve performance, information delays and the ways that attention may be paid to aggregate performance over dis-aggregated performance by sub-group suggest that a learning cycle may not be sufficient for equitably improving health outcomes. This suggests that an assumption of a feedback structure may be more useful as a tool for design for the future, rather than an analytic lens through which to assess current policy options.

\subsubsection*{Operationalizing trust and historical trauma} In the process of developing our feedback theory, it became apparent that some type of intangible stock such as trust would be needed to describe the mechanism through which the performance of algorithms feed back into health seeking behavior such as participating in screening, accepting treatment, or seeking second opinions after a questionable diagnosis. One approach to building this feedback structure might have been to present some aggregate ``Trust'' stock with a rough estimate of an adjustment time, relying on a graphical function to describe a non-linear relationship between algorithm performance and trust.  However, the design of this project-based learning cycle was to tightly integrate the model conceptualization with folks with professional, academic, and lived experience on these very topics. Community call discussions presented dynamics of accumulation, using a bathtub metaphor as a learning tool. The discussion quickly turned to the implications of this accumulation concept to think about ideas of historical trauma and conceptions of control for the management of individual and community inflows and outflows.  Building on these conversation, the modeling team developed a set of structures that reflected the nuance and priority of community participants.  (Dis)Trust could instead be thought of as a function of individual and collective memory of positive or negative experiences related to treatment outcomes, adjusting as the consequences of trust on health seeking behavior feed back to aggregate health outcomes.  Trust also is adjusted by an additional biflow representing the multiple other accumulations of harmful or positive experiences that White and Black Americans differentially accumulate in US healthcare interactions.

Additionally, initial conditions and time constants related to memory and trust were differentiated to reflect the historical injustices of deaths related to research studies like Tuskegee \cite{aslan2016tuskegee,scharff2010tuskegee}, case studies like Henrietta Lacks \cite{skloot2010,beskow2016hela} , and other negative experiences with healthcare systems of Black Americans. The group discussed this issue at length and agreed that the deaths and other harms experienced by Black Americans did not impact all Americans' trust in medical care. Put another way, White Americans equally aware of Tuskegee would not consider Tuskegee in their calculus of trust in medical care. To reflect these experiences, the stocks of trust and memory were also arrayed, with Black Americans beginning with a higher memory of death and lower trust in medical care than White Americans.

The purpose of presenting this structure, and discussing the genesis of this specific formulation of trust, is not intended to suggest that this specific formulation represents a standard for how trust should be modeled in future work. The model behavior is certainly sensitive to modeling choices of mechanisms of trust.  Instead, we believe that this formulation, by taking seriously the mechanisms of trust accrual and the designation of initial conditions and rates of decay, might guide future modelers to take a community-driven approach to these formulations as well.  

\subsection*{Process insights}

\subsubsection*{System dynamics as a language to talk about structural racism, inequity and bias} 

Racial bias is a singular terms that encompasses a complex set of concepts. From the system dynamics perspective, racial bias can include differences in the information cues used to make (in this case) medical decisions, differences in how the same cues are interpreted for different groups or individuals, differences in key parameters (like availability of medical treatment) that systematically disadvantage some groups, and differences in the initial conditions of state variable that can then create path dependent effects on health outcomes of interest, to name a few. The processes described in this paper shed light on how the tools of system dynamics can support the precise and formal representation of bias, which allows for a deeper discussions and understanding of how they are interrelated.  While some have included racism as a variable in their SD models \cite{frerichs2016}, our work adds to the discussion by using SD to explicitly reflect where racial bias manifests in institutional structures.  We intend that this model will be a starting point of future work to delineate multiple ways to conceptualize, describe, model, and, most importantly, work to dismantle structural racial bias.

\subsubsection*{Feasibility of replicating this process for new topics}
The modeling process described in this paper was one that evolved with the calls and the engagement of call participants.  The proposition and potential of using virtual group model building to build capacity for using SD through a specific project (e.g., racial disparities health, data, AI/ML technologies) was unknown at the start, but the outcomes suggest that parts of the process have potential for replication.   Specifically, the facilitated community calls appeared useful for broader model-based discussions and model review, and a 90 minute call and structure, which provided time for participants to introduce themselves and for facilitators to recap the overall goals of the group and work from previous calls, created conditions for participants to be able to engage meaningfully, even if they could not attend each call.  While video conferencing enables more people to participate, the relatively small number of attendees (2-6 in addition to the facilitators) promoted more discussion than is typically feasible with larger video conferencing groups.  A higher number of participants may have negatively impacted depth of discussion and trust built among participants via video conferencing. 

Additionally, the virtual modeling done between each community call between the group of call facilitators was an innovation of the process, which had positive impacts in terms of developing modeling capabilities and model insights among the team. In the facilitators' previous group model building experiences, even when in-person meetings were a possibility, the bulk of the modeling has more frequently been done by one or more modelers taking turns working on the model with some model check-in meetings. In this project, after the initial perceived success of the virtual modeling, the majority of the rest of the modeling was done with all three facilitators present.  In this process, formulation of rate equations, selection of stocks and initial parameters were all undertaken in real-time, allowing for deeper conversations on the implications of one formulation over another, and more rapid iteration than is typically possible in group model building work. The substantive area and modeling expertise of each facilitator was then able to be leveraged and increased in each model iteration, building structural insight from the model and further ownership of the model and process as the call facilitators engaged the larger community in monthly calls. 

\section*{Conclusion}
The field of ML/AI-supported healthcare technologies is rapidly developing, but with the risk of automating historical and structural racial biases.  The practice of system dynamics, particularly when undertaken through the values and lens of Community Based System Dynamics, has a tremendous potential to facilitate a deeper understanding of the implications of these technologies and the biases in available data while creating communities of practice equipped with tools to prevent these outcomes.  In the context of racial bias in ML/AI-based supportive technologies, where engaging a diverse community of Black and brown folk (and allies) working across different sectors in different capacities is essential, using this short-term learning approach that facilitates shared learning may be a beneficial way of striking a balance of providing a clearly structured context for learning SD tools and individual benefit of taking learned skills back to community and movement contexts relevant to them. While the rapid modeling work described in this paper focused principally on the structural racial disparities impacting the availability and collection of data, future work could more rigorously interrogate the decision rules and proxy variables underlying the algorithms themselves, and open up for larger discussion the health problem(s) that ML/AI-based products are being designed to better address. 

\section*{Acknowledgements}
The authors of this paper would like to thank the participants of the the SD for Racial Justice calls, whose participation drove the modeling process and insights discussed in this paper: Richard Carder, Haliday Douglas,  Nakisa Glover, Melanie Houston, Temple Lovelace, Anna Minsky, Cherish Molezion, Georgia Simposon and Zakiyah Shaakir-Ansari.  We are also grateful for the participants of the SD Learning Lab in 2018,  many of whom also designed and facilitated SD workshops at Data 4 Black Lives II in 2019: Mary Baxter, Haliday Douglas, Philip McHarris, Bella Roseman, Jamaal Sebastian-Barnes, Zakiyah Shaakir-Ansari, Jamila Smith-Loud, and LaTonya Wallace. We thank the organizers of the Data 4 Black Lives II who created space for this work at D4BL II: Yeshimabeit Milner, Lucas Mason-Brown and Max Clermont. Lastly, we thank the reviewers of this paper for their thoughtful comments and feedback.

\printbibliography

\end{document}